\documentclass[a4paper]{article}
\usepackage{latexsym}
\usepackage{sss}

\usepackage{graphicx} 
\usepackage{epsfig} 
\usepackage{amsmath,amssymb}

\newcommand{\real}{\mathbb{R}}
\newcommand{\E}{\mathbb{E}}
\newcommand{\PP}{\mathbb{P}}
\newcommand{\re}{\mathrm{e}}
\newenvironment{Proof}{\removelastskip\par\medskip \noindent{\em Proof.} \rm}{\penalty-20\null\hfill$\square$\par\medbreak}

\usepackage[small]{authorindex}

\newtheorem{prop}{Proposition}[section]
\newtheorem{lemma}[prop]{Lemma}

\usepackage{float}
\usepackage{caption} 
\usepackage{subcaption} 

\usepackage{xcolor}
\definecolor{ForestGreen}{RGB}{34,139,34}
\definecolor{mauve}{rgb}{0.7,0,0.43}
\definecolor{dkgreen}{rgb}{0,0.6,0}
\definecolor{darkgreen}{rgb}{0,0.6,0}
\definecolor{darkorange}{rgb}{1.0, 0.55, 0.0}
\definecolor{lightblue}{rgb}{0,0.2,0.5}
\definecolor{blue1}{rgb}{0,0.1,0.9}

\definecolor{lightblue}{rgb}{0,0.2,0.5}

\usepackage[colorlinks=true, urlcolor=blue,linkcolor=blue, citecolor=lightblue]{hyperref}

\usepackage{textcomp}

\usepackage{accsupp}    

\newcommand{\noncopynumber}[1]{
    \BeginAccSupp{method=escape,ActualText={}}
    #1
    \EndAccSupp{}
}
  	
\usepackage{listings}
\lstdefinelanguage{Maple}{
    morekeywords={proc, if, return, map, op, int, for, do, local, nops, convert, end},
    sensitive=false, 
    morecomment=[l]{//}, 
    morecomment=[s]{/*}{*/}, 
    morestring=[b]" 
} 

\usepackage{flushend} 

\begin{document}
\date{}
\title{\LARGE \textbf{
 An Algorithm for the Computation of Joint Hawkes Moments with Exponential Kernel\footnote{Submitted to the Proceedings of the 53rd International Symposium on Stochastic Systems Theory and Its Applications (SSS'21).}
}
}
\author{Nicolas Privault
\\ 
\small
School of Physical and Mathematical Sciences, 
\\
\small Nanyang Technological University 
\\ 
\small 
21 Nanyang Link, Singapore 637371
\\
\small
E-mail: \texttt{nprivault@ntu.edu.sg}
}

\maketitle
\thispagestyle{empty}
\abstract{
The purpose of this paper is to present a recursive algorithm and its implementation in Maple and Mathematica for the computation of joint moments and cumulants of Hawkes processes with exponential kernels. Numerical results and computation times are also discussed. Obtaining closed form expressions can be computationally intensive, as joint fifth cumulant and moment formulas can be respectively expanded into up to 3,288 and 27,116 summands. 
} 
 

\section{Introduction}
Hawkes processes \cite{hawkes1971} are self-exciting point processes
that have found applications in fields such 
as neuroscience \cite{cardanobile},
 genomics analysis \cite{reynaud-bouret},
as well as finance \cite{embrechts2}, or social media \cite{younglee}. 

\medskip 

The analysis of statistical properties of Hawkes processes is made
difficult by their recursive nature, making the computation of moments
difficult. 
 In \cite{jovanovic}, a tree-based method for the computation of cumulants 
 has been introduced, with an explicit computation of third order cumulants.
 Ordinary differential equation (ODE) methods have been applied in 
 \cite{dassios-zhao2} to the computation of
 the moment and probability generating functions of
 (generalized) Hawkes processes and their intensity, with the computation
 of first and second moments in the stationary case, see also
 \cite{errais}, and \cite{cui} and \cite{daw} for other ODE-based
 approaches. 

 \medskip

 In \cite{bacry}, stochastic calculus and martingale arguments have been
 applied to the computation of first and second order moments,
 however those approaches seem difficult to generalize to higher-orders
 moments.
 In \cite{vargas},
 cumulant recursion formulas have been obtained
 for general random variables
 using martingale brackets. 
 Third-order cumulant expressions for Hawkes processes
 have been used in \cite{achab} for the
 the analysis of order books in finance,
 and in
\cite{ocker},
\cite{montangie} 
for neuronal networks.

\medskip

In \cite{hawkescumulants},
the cumulants of Hawkes processes have been computed
using using Bell polynomials, based on a recursive relation for
the Probability Generating Functional (PGFl) of self-exciting
point processes started from a single point. 
 This provides a closed-form alternative to the tree-based approach of \cite{jovanovic}. 

\medskip 

In this note we apply the algorithm of \cite{hawkescumulants} 
to the recursive computation of joint moments of all orders
of Hawkes processes, and present the
corresponding codes written in Maple and Mathematica.
The algorithm uses sums over partitions and Bell polynomials to compute
 joint cumulants in the case of an 
 exponential branching intensity on $[0,\infty )$.

\medskip

We proceed as follows.
 After reviewing some combinatorial identities in Section~\ref{s2}, 
 we will consider the computation of the 
 joint cumulants of self-exciting Hawkes Poisson cluster processes
 in Section~\ref{s3}. 
 Explicit computations for the time-dependent 
 joint third and fourth cumulants 
 of Hawkes processes with exponential kernels are presented
 in Section~\ref{s4},
 and are confirmed by Monte Carlo estimates. 
 \section{Joint moments and cumulants}
 \label{s2} 
 In this section we present background combinatorial results 
 that will be needed in the sequel. 
 Given the Moment Generating Function (MGF) 
\aimention{Thiele, T.N.}
\begin{align} 
\nonumber 
 & 
  M_X (t_1,\ldots , t_n) : = 
 \E \big[ \re^{t_1X_1+\cdots + t_n X_n} \big] 
\\
\nonumber
 & \qquad = 
 1 + \sum_{k_1,\ldots , k_n \geq 1} 
 \frac{t^{k_1}_1\cdots t^{k_n}_n}{n!} 
 \E [ X^{k_1}_1 \cdots X^{k_n}_n ], 
\end{align} 
 of a random vector $X=(X_1,\ldots , X_n)$,
 the joint {cumulants} of $(X_1,\ldots , X_n)$ of orders
 $(l_1,\ldots , l_n)$ are the coefficients 
 $\kappa_{l_1,\ldots , l_n} (X)$ appearing in the log-MGF expansion 
\begin{align} 
\label{cgf} 
 & 
\hskip-0.4cm
\log M_X (t_1,\ldots , t_n)
 =
\log \big( \E\big[\re^{t_1X_1+\cdots + t_n X_n}\big] \big) 
\\
\nonumber
 & = 
 \sum_{l_1,\ldots , l_n\geq 1} \frac{t^{l_1}_1\cdots t^{l_n}_n}{l_1! \cdots l_n!}
 \kappa_{l_1,\ldots , l_n} (X_1,\ldots , X_n), 
\end{align} 
for $(t_1,\ldots , t_n)$ in a neighborhood of zero in $\real^n$.
 In the sequel we let
 $$
 \kappa (X_1,\ldots , X_n)
 := \kappa_{1,\ldots , 1} (X_1,\ldots , X_n),
 \quad
 n\geq 1, 
$$ 
 and
 $$
 \kappa^{(n)} (X)
 := \kappa_{1,\ldots , 1} (X,\ldots , X),
 \quad n\geq 1.
 $$ 
The joint moments of $(X_1,\ldots , X_n)$
are then given by the joint moment-cumulant relation
\begin{equation} 
\nonumber 
\E [ X_1 \cdots X_n ] 
 = 
 \sum_{l=1}^n
   \sum_{\pi_1\cup \cdots \cup \pi_l = \{1,\ldots , n\}}
  \prod_{j=1}^l 
  \kappa^{(|\pi_j|)} \big( (X_i)_{i \in \pi_j} \big).
\end{equation} 
 where the sum runs over the partitions 
 $\pi_1,\ldots , \pi_k$ of the set $\{ 1 , \ldots , n \}$,
 By the multivariate Fa\`a di Bruno formula, \eqref{cgf} can be inverted as
\begin{align*} 
 &     \kappa (X_1,\ldots , X_n) 
  \\
   & = 
 \sum_{l=1}^n
  (l-1)!
  (-1)^{l-1}
 \hskip-0.1cm
 \sum_{\pi_1\cup \cdots \cup \pi_l = \{1,\ldots , n\}}
  \prod_{j=1}^l 
  \E \Bigg[ \prod_{i\in \pi_j} X_i \Bigg]. 
\end{align*} 
 In the univariate case, the moments $\E[X^n]$ of a random variable $X$ 
 are linked to its cumulants $\big(\kappa^{(n)}(X)\big)_{n\geq 1}$
 through the relation 
\begin{align} 
\nonumber 
 \E [ X^n ] 
 & = 
 B_n \big( \kappa^{(1)}(X), \ldots , \kappa^{(n)} (X) \big) 
 \\
 \nonumber
 & = 
 \sum_{k=1}^n 
 B_{n,k} ( \kappa^{(1)} (X) \cdots \kappa^{(n-k+1)} (X) ), 
\end{align} 
where 
\begin{align} 
 \nonumber
 & 
 \hskip-0.3cm
 B_{n,k} ( a_1 , \ldots , a_{n-k+1} ) 
 = 
 \frac{n!}{k!} 
 \sum_{l_1+\cdots + l_k=n \atop 
 l_1\geq 1,\ldots ,l_k \geq 1 
 } 
 \frac{a_{l_1}}{l_1!} 
 \cdots 
 \frac{a_{l_k}}{l_k!}
  \\
  \label{dfjkl0}
 & = 
 \sum_{\pi_1 \cup \cdots \cup \pi_k = \{ 1, \ldots , n \}} 
 a_{|\pi_1|}(X) \cdots a_{|\pi_k|}(X), 
\end{align} 
 $1\leq k \leq n$, is the partial Bell polynomial 
 of order $(n,k)$,
 where
 the sum \eqref{dfjkl0} holds on the integer compositions
 $(l_1,\ldots ,l_k)$ of $n$,
 see e.g. Relation~(2.5) in \cite{elukacs},
 and 
\begin{align*} 
 B_n ( a_1 , \ldots , a_n ) 
 & = 
 \sum_{k=1}^n
 B_{n,k} ( a_1 , \ldots , a_{n-k+1} )
\end{align*} 
 is the complete Bell polynomial of degree $n \geq 1$. 
 We also have the inversion relation 
\begin{align} 
\nonumber 
& \kappa^{(n)} (X)
\\
\nonumber 
& = \sum_{k=0}^{n-1} 
 k! (-1)^k 
B_{n,k+1} \big(
\E \big[ X \big] , \E \big[ X^2 \big] , \ldots , \E \big[ X^{n-k} \big]
\big) 
\end{align} 
$n \geq 1$,
see e.g. Theorem~1 of \cite{elukacs}, 
 and also \cite{leonov}, 
 Relations~(2.8)-(2.9) in \cite{mccullagh}, or 
 Corollary~5.1.6 in \cite{stanley}. 

 \medskip

 As an example we consider the recursive computation of Borel cumulants as an example.
 Let $(X_n)_{n\geq 0}$ be a branching process started at $X_0=1$
 with Poisson distributed offspring count $N$ of parameter $\mu \in (0,1)$,
 and let $X$ denote the total count of offsprings generated by $(X_n)_{n\geq 0}$
 It is known,
 see \cite{polya-szego}
 and \S~3.2 of \cite{consul}
 that $X$ has the {Borel distribution} 
\index{Lagrangian distribution}
\index{distribution!Lagrangian}
\index{Borel distribution}
\index{distribution!Borel}
\aimention{Consul, P.C.}
\aimention{Famoye, F.}
\aimention{P{\'o}lya, G.} 
\aimention{Szeg{\"o}, G.} 
 $$
 \PP ( X = n )
 = \re^{-\mu n}\frac{(\mu n)^{n-1}}{n!},
 \qquad n \geq 1. 
$$
 We have $\kappa^{(1)} (X)= 1/(1-\mu)$ and the induction relation
\begin{align} 
 \nonumber 
& \kappa^{(n)}(X)
  = 
 \frac{\mu}{1-\mu}
 \big( B_n \big( \kappa^{(1)}(X), \ldots , \kappa^{(n)}(X) \big)
 - \kappa^{(n)}(X) \big)
 \\
 \label{fjls}
 & = 
 \frac{\mu}{1-\mu}
  \sum_{k=2}^n
 B_{n,k} \big( \kappa^{(1)}(X), \ldots , \kappa^{(n-k+1)}(X) \big) 
, 
\end{align} 
 $n\geq 2$, see \S~8.4.3 in \cite{consul}
 and Proposition~2.1 in \cite{hawkescumulants}. 
 The recursion \eqref{fjls} is implemented in the following Maple 
 code. 
 
\medskip

\begin{lstlisting}[language=Maple]
c := proc(n, mu) local tmp, k, z1; option remember; if n = 1 then return 1/(1 - mu); end if;
  tmp := 0; z1 := []; for k from n by -1 to 2 do z1 := [op(z1), c(n - k + 1, mu)]; tmp := tmp + IncompleteBellB(n, k, op(z1)); end do;
  return mu*tmp/(1 - mu); end proc;
m := proc(n, mu) local tmp, z, k; option remember; if n = 0 then return 1; end if;
  tmp := 0; z := []; for k from n by -1 to 1 do z := [op(z), c(n - k + 1, mu)]; tmp := tmp + IncompleteBellB(n, k, op(z)); end do;
  return tmp; end proc;
\end{lstlisting}

\vspace{-0.6cm}

 \noindent
 In particular, the command ${\rm c}(2,\mu)$ in Maple yields 
 the second cumulant $\kappa^{(2)}(X) = \mu / (1-\mu)^3$, 
 and by the commands ${\rm c}(3,\mu)$ and ${\rm c}(4,\mu)$ 
 we find 
$\kappa^{(3)}(X)
=
\mu ( 1+2\mu)/(1-\mu)^5$,
 and
 $
 \kappa^{(4)}(X)
 = \mu ( 1 + 8\mu + 6 \mu^2 )/(1-\mu)^7$, 
 see also (8.85) page 159 of \cite{consul}. 
Those results can be recovered from
the command ${\rm c}[2,\mu]$,
${\rm c}[3,\mu]$ and ${\rm c}[4,\mu]$ 
using the following Mathematica code. 

\medskip

\begin{lstlisting}[language=Mathematica]
c[n_, mu_] := c[n, mu] = (Module[{tmp, k}, If[n == 1, Return[1/(1 - mu)]]; tmp = 0; z1 = {}; 
  For[k = n, k >= 2, k--, z1 = Append[z1, Block[{i = n - k + 1}, c[i, mu]]]; tmp += BellY[n, k, z1]];
  Simplify[mu*tmp/(1 - mu)]]);
m[n_, mu_] := (Module[{tmp, z, k}, tmp = 0; If[n == 0, Return[1]]; 
   z = {}; For[k = n, k >= 1, k--, z = Append[z, Block[{i = n - k + 1}, c[i, mu]]]; tmp += BellY[n, k, z]]; Simplify[tmp]])
\end{lstlisting}

\vspace{-0.8cm}

\section{Joint Hawkes cumulants} 
\label{s3}
\begin{figure}[H]
  \centering
   \includegraphics[width=1\linewidth]{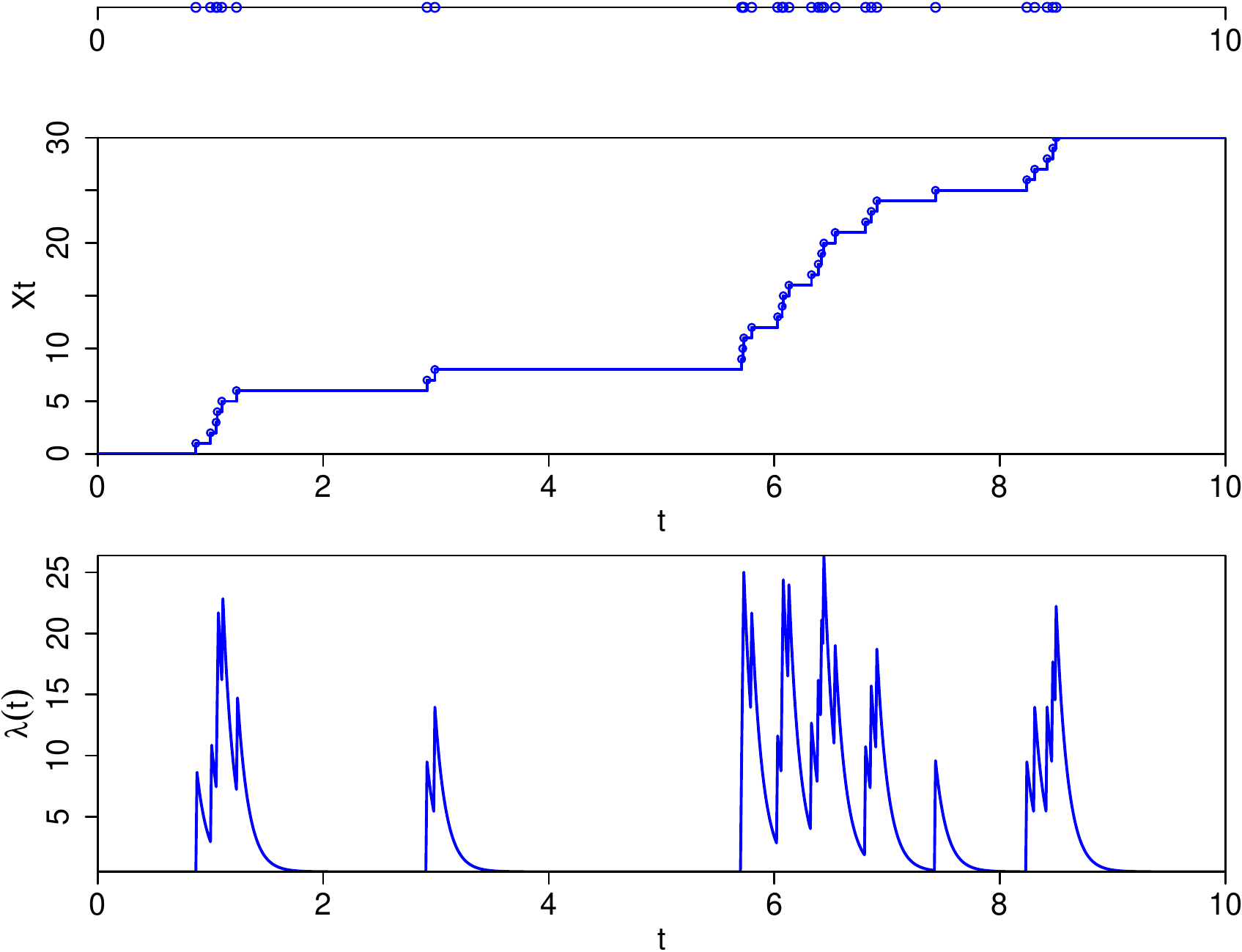} 
 \caption{Sample paths of $X_t$ and of the intensity $\lambda (t)$.} 
    \label{fig00}
\end{figure}

In the cluster process framework of \cite{hawkes}, 
we consider a real-valued self-exciting point process on $[0,\infty )$, 
with Poisson
offspring intensity $\gamma (dx)$ and Poisson
immigrant intensity $\nu (dx)$ on $[0,\infty )$, built on the space
\begin{eqnarray*}
  \lefteqn{
    \Omega = \big\{
 \xi = \{ x_i \}_{i\in I} \subset [0,\infty ) \ : \
  }
  \\
  & &
  \#( A \cap \xi ) < \infty 
 \mbox{ for all compact } A\subset [0,\infty )
 \big\}
 \end{eqnarray*} 
 of locally finite configurations on $[0,\infty )$, whose elements 
 $\xi \in \Omega$ are identified with the Radon point measures 
 $\displaystyle \xi (dz) = \sum_{x\in \xi} \epsilon_x (dz)$, 
 where $\epsilon_x$ denotes the Dirac measure at $x\in \real_+$. 
 In particular, any initial immigrant point $y \in \real_+$ branches into a Poisson
 random sample denoted by
 $\xi_\gamma (y + d z ) \displaystyle = \sum_{x\in \xi} \epsilon_{x+y} (d z )$
 and centered at $y$, with intensity measure $\gamma (y + d z )$ on $[0,\infty )$. 
 Figure~\ref{fig00} presents a graph of the point measure $\xi (dz)$ followed by
 the corresponding sample paths of the self-exciting counting process
 $\displaystyle X_t (\xi ) := \xi ( [0,t]) = \sum_{x\in \xi} {\bf 1}_{[0,t]}(x)$ 
 and its stochastic intensity $\lambda_t$, $t\in [0,10]$,
 in the exponential kernel example
 of the next section. 
 
 \medskip

\vskip-0.2cm

\noindent 
 In the sequel, we assume that $\gamma ( [0,\infty ) ) < 1$ and 
 consider the integral operator $\Gamma$ defined as
 $$
 \Gamma f (z) = \int_0^\infty f(z+y ) \gamma (dy), \qquad z\in \real_+, 
 $$
 and the inverse operator $(I_d-\Gamma)^{-1}$ given by 
 \begin{align*}
   & 
 (I_d-\Gamma)^{-1} f(z ) 
   = f(z) 
   \\
   &
  \quad + \sum_{m=1}^\infty \int_{\real_+^m} f(z+x_1+\cdots + x_m) \gamma ( dx_1)
  \cdots \gamma ( dx_m), 
\end{align*} 
 with
  \begin{align*}
   & 
 (I_d-\Gamma)^{-1} \Gamma f(z ) 
    =  (I_d-\Gamma)^{-1} f(z ) - f(z ) 
    \\
   &
  \quad = \sum_{m=1}^\infty \int_{\real_+^m} f(z+x_1+\cdots + x_m) \gamma ( dx_1)
  \cdots \gamma ( dx_m), 
\end{align*} 
$z\in \real_+$.
 The first cumulant
 $\kappa_z^{(1)}(f)$ of $\displaystyle \sum_{x\in \xi} f(x)$
 given that $\xi$ is started 
 from a single point at $z\in \real_+$ 
 is given by 
 $\kappa_z^{(1)}(f)  = (I_d-\Gamma )^{-1} f(z)$ for $n=1$. 
The next proposition provides a way to compute the
higher order cumulants $\kappa_z^{(n)}(f)$ of $\displaystyle \sum_{x\in \xi} f(x)$
 given that $\xi$ is started 
 from a single point at $z\in \real_+$ 
 by an induction relation based on set partitions,
 see Proposition~3.5 in \cite{hawkescumulants}. 
\begin{prop}
\label{djklds}
 For $n\geq 2$, the joint cumulants 
 $\kappa_z^{(n)}(f_1,\ldots , f_n)$ 
 of $\displaystyle \sum_{x\in \xi} f_1(x), \ldots , \sum_{x\in \xi} f_n(x)$
 given that $\xi$ is started 
 from a single point at $z\in \real_+$ 
 are given by the induction relation 
\begin{align} 
  \label{fjkl}
   & \kappa_z^{(n)}(f_1,\ldots , f_n) 
 \\
 \nonumber
 &
 = 
 \sum_{k=2}^n
 \sum_{\pi_1\cup \cdots \cup \pi_k = \{1,\ldots , n\}} 
 (I_d-\Gamma )^{-1} \Gamma \kappa_z^{(|\pi_j|)} ((f_i)_{i\in \pi_j})
,
\end{align} 
$n \geq 2$, where the above sum is over set partitions
$\pi_1\cup \cdots \cup \pi_k=\{1,\ldots , n\}$, $k=2,\ldots , n$,
and $|\pi_i|$ denotes the cardinality of the set $\pi_i \subset \{1,\ldots , n\}$.
\end{prop}

\noindent
The joint cumulants
 $\kappa^{(n)} (f_1,\ldots , f_n)$ 
of $\displaystyle \sum_{x\in \xi} f_1(x), \ldots , \sum_{x\in \xi} f_n(x)$
 can be obtained as a consequence
 of Proposition~\ref{djklds}, by the combinatorial summation 
 \begin{align}
   \label{al} 
 & 
  \kappa^{(n)} (f_1,\ldots , f_n)
  \\
  \nonumber
  & = 
  \sum_{k=1}^n
 \sum_{\pi_1\cup \cdots \cup \pi_k = \{1,\ldots , n\}} 
 \int_0^\infty \prod_{j=1}^k
 \kappa_z^{(|\pi_j|)}((f_i)_{i\in \pi_j})
 \nu ( dz ), 
\end{align}
 see Corollary~3.4 and Proposition~3.5 in \cite{hawkescumulants}. 
 Joint moments can  then be recovered by the joint moment-cumulant relation
 \begin{align} 
   \label{mc}
   &  \hskip-0.3cm
   \E \Bigg[
     \sum_{x\in \xi} f_1(x)
     \cdots
     \sum_{x\in \xi} f_n(x)
     \Bigg] 
     \\
   \nonumber
   & = 
 \sum_{l=1}^n
   \sum_{\pi_1\cup \cdots \cup \pi_l = \{1,\ldots , n\}}
  \prod_{j=1}^l 
  \kappa^{(|\pi_j|)} ((f_i)_{i\in \pi_j}), 
\end{align} 
  which can be inverted as 
\begin{align*} 
& \kappa^{(n)} (f_1,\ldots ,  f_n )
  \\
  & \hskip-0.05cm
   = 
 \sum_{l=1}^n
  (l-1)!
  (-1)^{l-1}
 \hskip-0.5cm
 \sum_{\pi_1\cup \cdots \cup \pi_l = \{1,\ldots , n\}}
  \prod_{j=1}^l 
    \E \Bigg[
    \prod_{i\in \pi_j} \sum_{x\in \xi} f_i(x)
     \Bigg]. 
\end{align*} 

\section{Joint Hawkes moments with exponential kernel} 
\label{s4}
\noindent 
In this section we consider the exponential kernel
$\gamma (dx) = a {\bf 1}_{[0,\infty )} (x) \re^{-bx} dx$, $0< a < b$, 
and constant Poisson intensity $\nu (dz) = \nu dz$, $\nu >0$. In this case,
$$
\displaystyle X_t (\xi ) := \xi ( [0,t]) = \sum_{x\in \xi} {\bf 1}_{[0,t]}(x),
 \quad t\in \real_+, 
$$ 
defines the self-exciting Hawkes process with stochastic intensity
$$ 
 \lambda_t := \nu + a \int_0^t \re^{-b(t-s)} dX_s, \qquad t\in \real_+.
$$
 In this case, the integral operator $\Gamma$ satisfies 
 $$
 \Gamma f (z) = a \int_0^\infty f(z+y ) \re^{-by} dy, \quad z\in \real_+, 
 $$
 and the recursive calculation of joint moments and cumulants
 will be performed by evaluating
 $(I_d-\Gamma )^{-1} \Gamma$
 in Proposition~\ref{djklds}
 on the family of functions $e_{p,\eta,t}$ of the form 
 $e_{p,\eta, t}(x) := x^p \re^{\eta x} {\bf 1}_{[0,t]}(x)$,
 $\eta < b$, $p\geq 0$,
 as in the next lemma.
 \begin{lemma}
   \label{l1} 
   For $f$ in the linear span generated by the functions
   $e_{p,\eta, t}$, $p\geq 0$, $\eta \in \real$,
   the operator 
   $( I_d - \Gamma )^{-1} \Gamma$ is given by
$$ 
 ( I_d - \Gamma )^{-1} \Gamma 
 f ( z ) 
 = 
 a \int_0^{t-z}
 f(z+y) \re^{( a -b)y}
 dy, 
$$ 
 $z\in [0,t]$.
 \end{lemma}
 \begin{Proof}
   For all $p,\eta \geq 0$ we have the equality 
     \begin{align*} 
& ( I_d - \Gamma )^{-1} \Gamma 
 e_{p,\eta, t} ( z ) 
 \\
  & = 
\sum_{n=1}^\infty
 \int_{[0,t]^n} 
 e_{p,\eta, t} ( z+x_1+\cdots + x_n )
 \gamma ( dx_1 ) \cdots \gamma ( dx_n )  
 \\
  & = 
 \sum_{n=1}^\infty
\frac{a^n}{(n-1)!}
 \int_0^{t-z}
 (z+y)^p
 \re^{\eta (z+y)}
 y^{n-1} \re^{-by}
 dy
 \\
  & = 
 a 
 \re^{\eta z} \int_0^{t-z}
 (z+y)^p \re^{(\eta + a -b)y}
 dy, \quad z\in [0,t],  
\end{align*} 
     which follows from the fact that the sum $\tau_1+\cdots +\tau_n$
     of $n$ exponential random variables with parameter $b>0$ has
     a gamma distribution with shape parameter $n\geq 1$ and scaling parameter
     $b>0$.
\end{Proof}
 Using Lemma~\ref{l1}, we can rewrite \eqref{fjkl}
 for $t_1<\cdots < t_n$ as 
 \begin{align}
   \label{fdsf} 
 & \kappa_z^{(n)}\big({\bf 1}_{[0,t_1]},\ldots , {\bf 1}_{[0,t_n]}\big) 
 \\
 \nonumber
 &
 \hskip-0.1cm
 = 
 \sum_{k=2}^n
 \sum_{\pi_1\cup \cdots \cup \pi_k = \{1,\ldots , n\}} 
  \hskip-0.1cm
 \int_0^{t_1-z}
  \hskip-0.5cm
  a
  \re^{( a -b)y}
 \kappa_{z+y}^{(|\pi_j|)} \big( \big( {\bf 1}_{[0,t_i]}\big)_{i\in \pi_j} \big)
 dy, 
\end{align} 
 with
\begin{align*} 
\kappa_z^{(1)}({\bf 1}_{[0,t]}) & = (I_d-\Gamma )^{-1} {\bf 1}_{[0,t]}(z)
\\
 & =  
\frac{b}{b-a} +\frac{a}{a-b} \re^{(a-b)(t-z)},
\quad z\in [0,t] ,
\end{align*} 
 if $a\not= b$, and
$$ 
\kappa_z^{(1)}({\bf 1}_{[0,t]}) = (I_d-\Gamma )^{-1} {\bf 1}_{[0,t]}(z)
 = 1 + a (t-z), 
$$ 
 $z\in [0,t]$, if $a=b$.
The recursive computation of
$\kappa_z^{(n)}\big({\bf 1}_{[0,t_1]},\ldots , {\bf 1}_{[0,t_n]}\big)$
in \eqref{fdsf} is implemented in the following Mathematica
code using Lemma~\ref{l1}. 

\medskip

\noindent
The computation of joint Hawkes cumulants by
the recursive relation \eqref{al} is then implemented in the following code.

\medskip

\begin{figure}[H]
  \centering
    \includegraphics[width=1\linewidth]{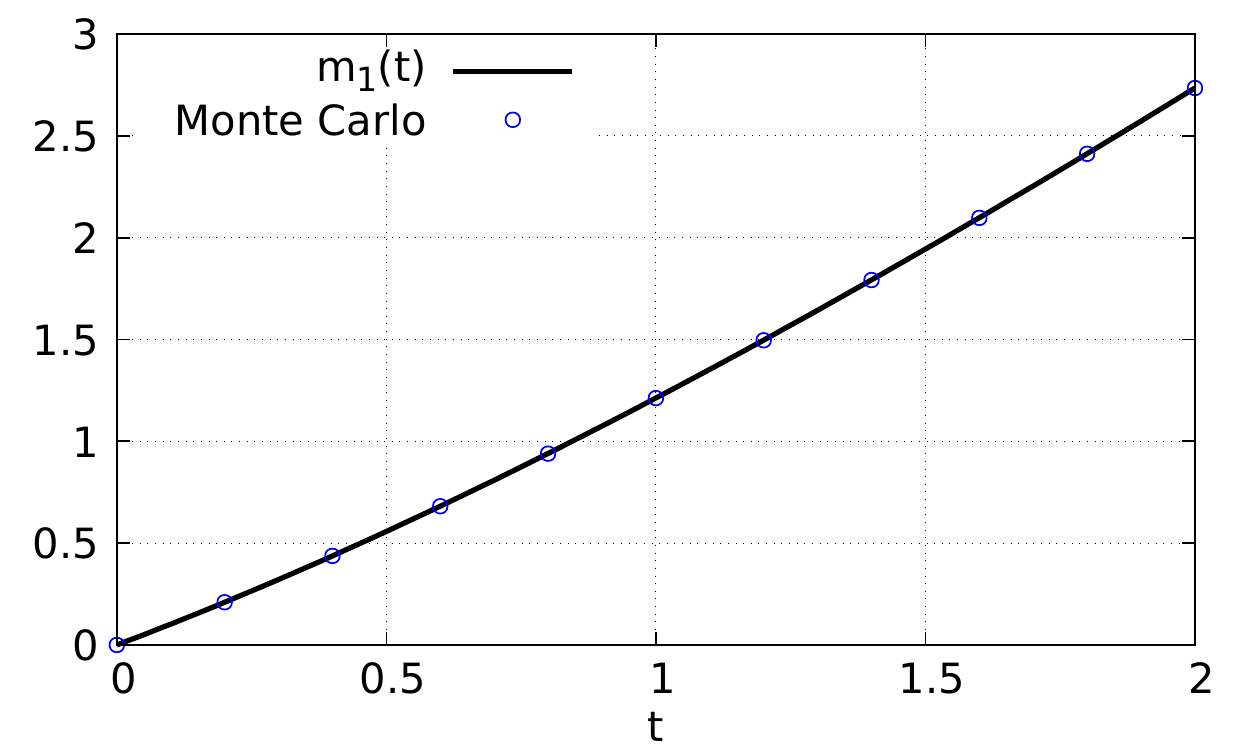} 
    \caption{First moment.} 
    \label{fig0}
\end{figure}

\begin{figure}[H] 
  \centering
      \includegraphics[width=1\linewidth]{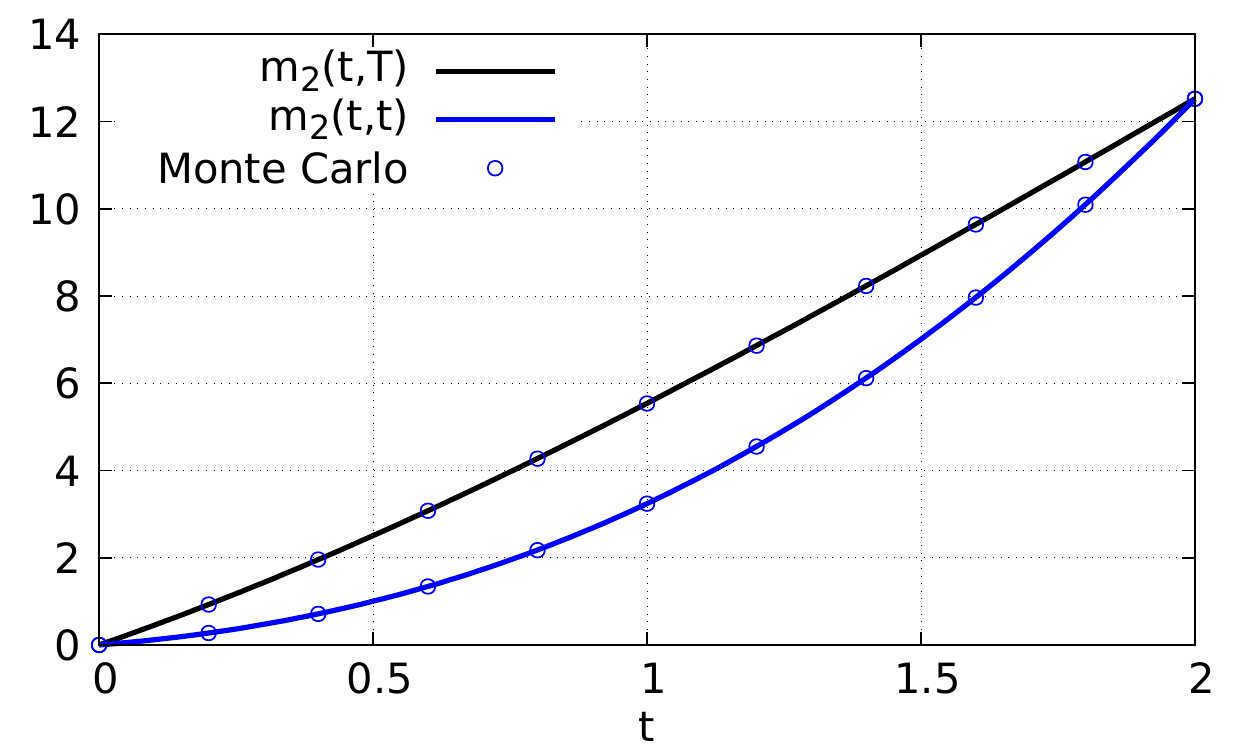} 
    \caption{Second joint moment.} 
 \label{fig1} 
\end{figure}

\begin{figure}[H]
  \centering
    \includegraphics[width=1\linewidth]{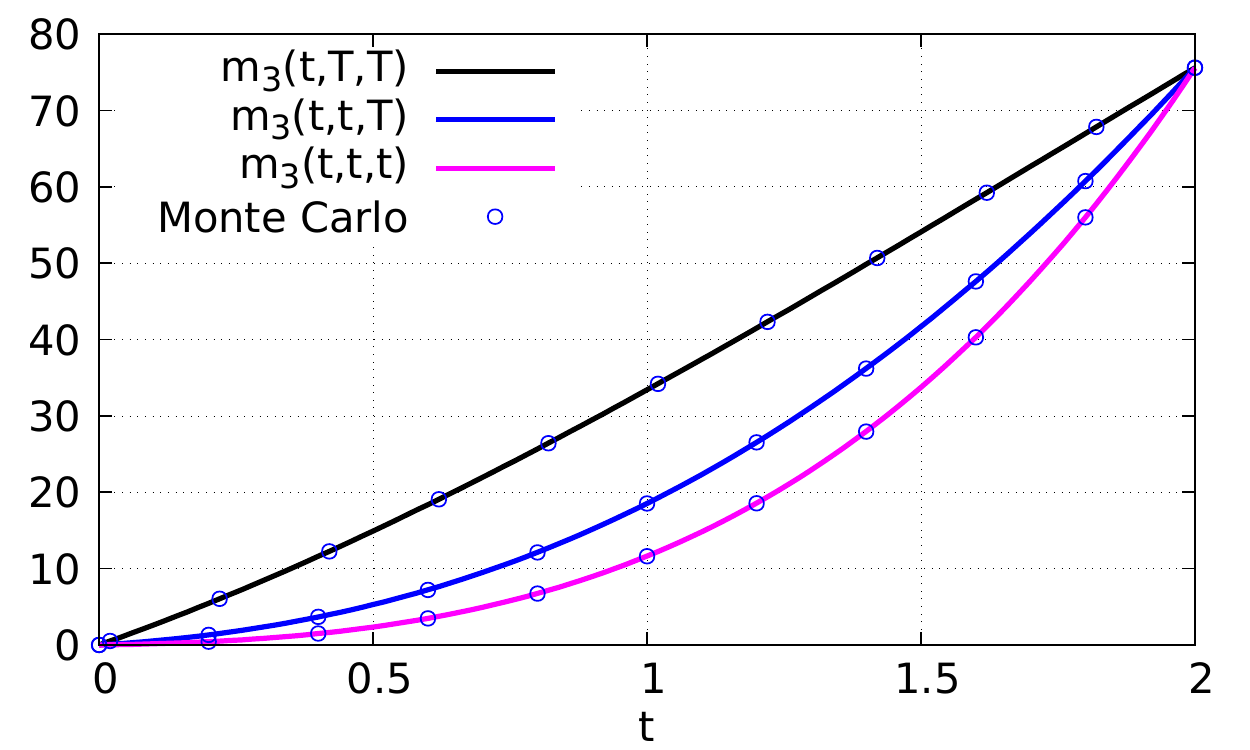} 
    \caption{Third joint moments.} 
      \label{fig2-a}
\end{figure}

\begin{figure}[H]
  \centering
      \includegraphics[width=1\linewidth]{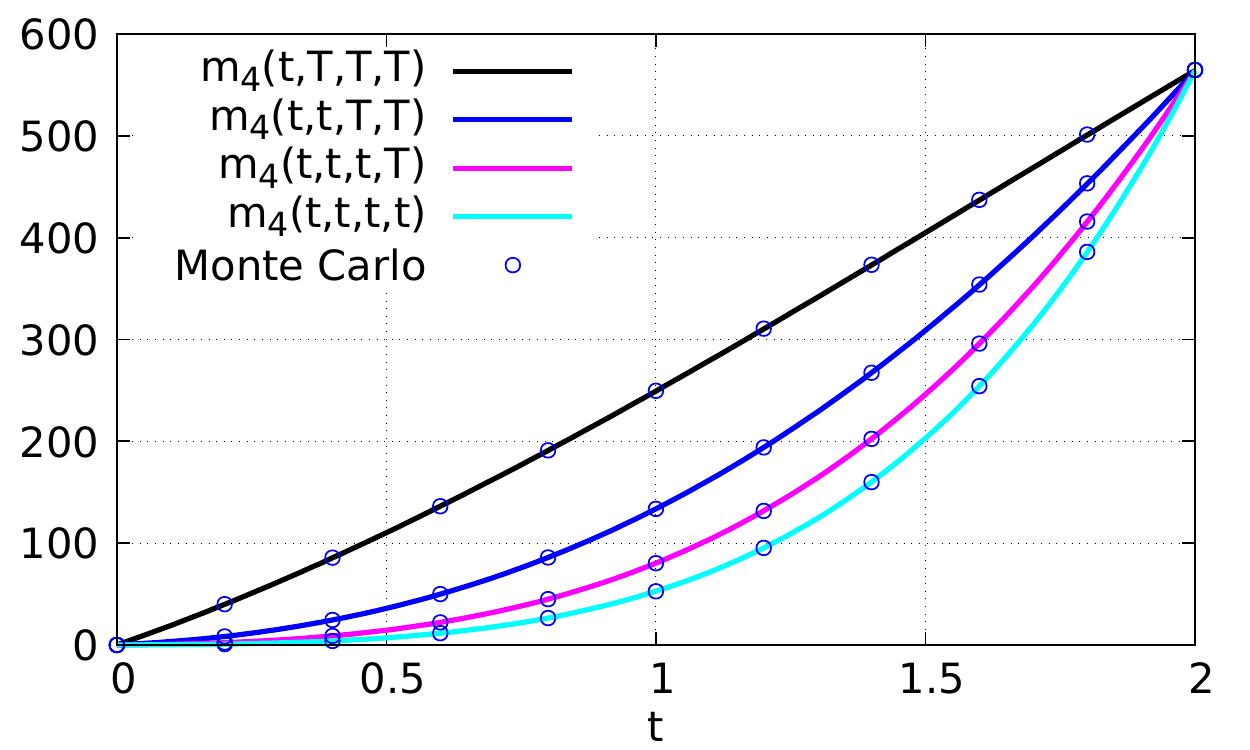} 
 \caption{Fourth joint moments.} 
       \label{fig2-b}
\end{figure}

\begin{figure}[H]
  \centering
    \includegraphics[width=1\linewidth]{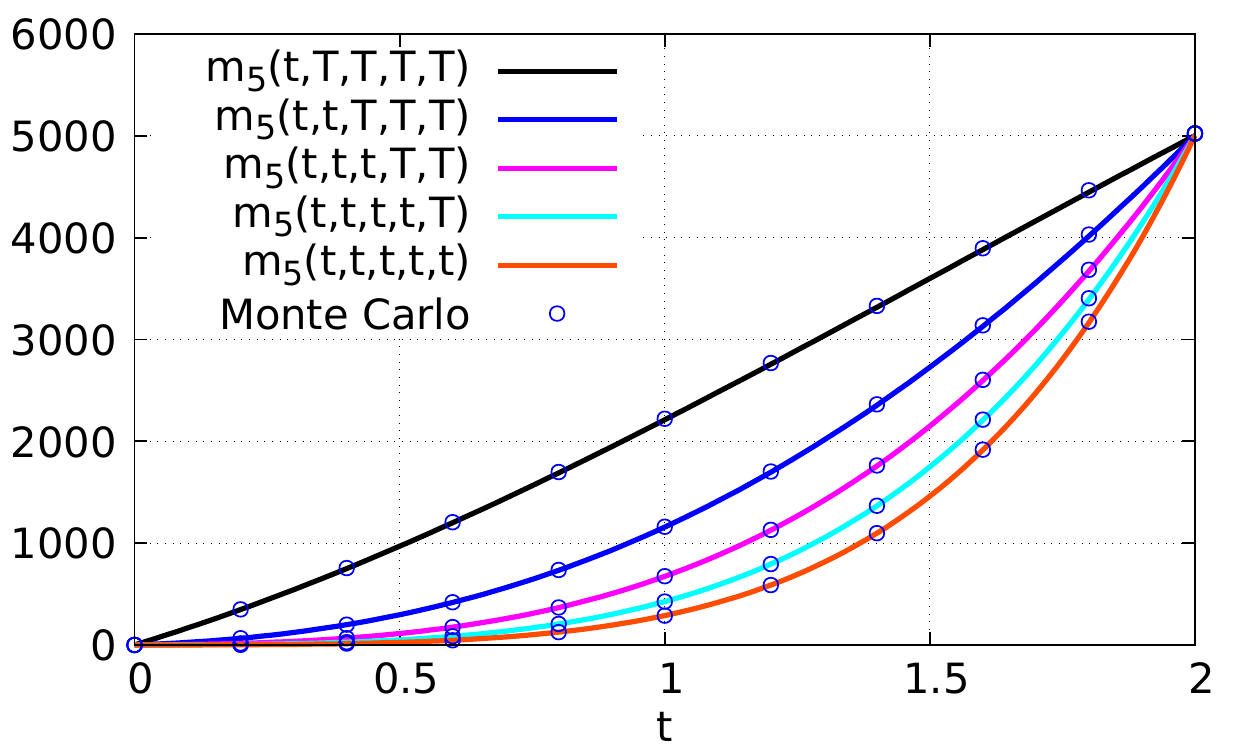} 
    \caption{Fifth joint moments.} 
      \label{fig3-a}
\end{figure}

\begin{figure}[H]
  \centering
     \includegraphics[width=1\linewidth]{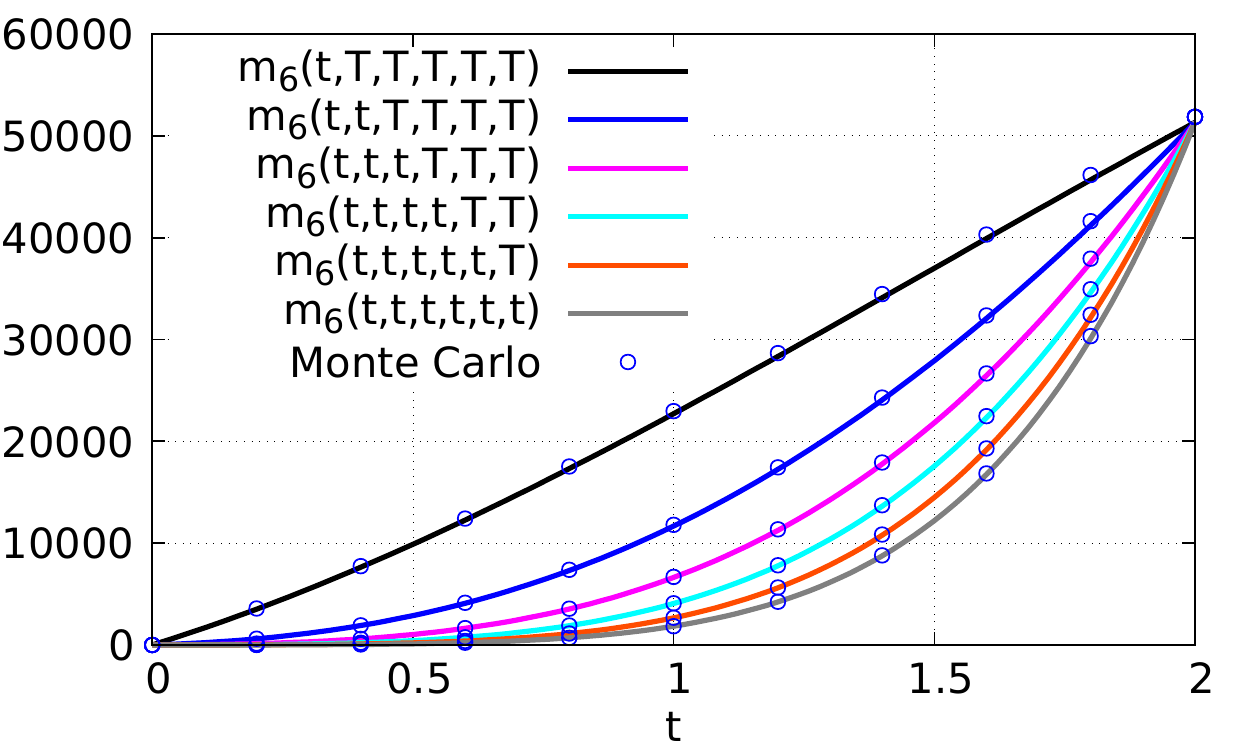} 
 \caption{Sixth joint moments.} 
       \label{fig3-b}
\end{figure}

\noindent
Finally, joint moments are computed from the joint moment-cumulant relation \eqref{mc}
which is implemented in the following code. 
The joint moments $\E[ X_{t_1} \cdots X_{t_n} ]$ 
of $X_{t_1}, \ldots , X_{t_n}$ are
obtained from the above code using the command
${\rm m}(a,b,[t_1,\ldots , t_n])$ in Maple or 
${\rm m}[a,b,\{t_1,\ldots , t_n\}]$ in Mathematica. 
 Figures~\ref{fig0} to \ref{fig3-b} 
 are plotted with $\nu=1$, $a=0.5$, $b=1$, $T=2$, and one million 
 Monte Carlo samples, and 
 Figure~\ref{fig0} presents the first moment
 $m_1(t) = {\rm m}(a,b,[t]) = {\rm m}[a,b,\{t\}]$. 

\vskip-0.2cm

\noindent 
 For example, the second joint moment of $(X_t,X_T)$ is obtained by the command
 $m_2(t,T) = {\rm m}(a,b,[t,T])$ in Maple or 
 $m_2(t,T) = {\rm m}[a,b,\{t,T\}]$ in Mathematica
 presented in appendix,
 which yields 
\begin{align*} 
& \E [ X_t X_T ] =
  \\ &
  \frac{1}{(a - b)^4}
\left(\frac{1}{2}
\re^{-a t - 
  b (t + T)} (2 b^4 t \re^{a t + b (t + T)} 
\right.
\\ &
 +      a^2 b (-\re^{a (2 t + T)} + \re^{2 b t + a T} + 
     2 b t \re^{2 a t + b T} 
     \\ & 
     + 2 b t \re^{b t + a (t + T)} ) + 
     2 a^3 (\re^{a (2 t + T)} - b t \re^{2 a t + b T} 
       \\ & 
   - 
        \re^{b t + a (t + T)} (1 + b t)) - 
        2 a b^2 (\re^{2 b t + a T} - 2 \re^{2 a t + b T}
          \\ & 
   - \re^{
        b t + a (t + T)} + 
   \re^{a t + b (t + T)} (2 + b t))) 
     \\ & 
  + (b^2 t + 
  a ( \re^{(a - b) t} - b t -1))
       \\ & 
\left.
\quad \times (b^2 T + 
     a ( \re^{(a - b) T} - b T -1))\right), 
   \end{align*} 
 see Figure~\ref{fig1}.

\vskip-0.2cm

\noindent 
Figures~\ref{fig2-a} and \ref{fig2-b}
show the numerical evaluation of third and fourth joint moments,
obtained from
$m_3(t_1,t_2,t_3) = {\rm m}(a,b,[t_1,t_2,t_3])= {\rm m}[a,b,\{t_1,t_2,t_3\}]$
and 
$m_4(t_1,t_2,t_3,t_4) ={\rm m}(a,b,[t_1,t_2,t_3,t_4])={\rm m}[a,b,\{t_1,t_2,t_3,t_4\}]$. 

\vskip-0.2cm

\vskip-0.2cm

\noindent 
The following tables count the (approximate) numbers of summands
appearing in joint cumulant and moment expressions when expanded as
a sum of the form 
$$
 \sum_{k,l,p_1,\ldots, p_n \geq 0\atop
   q_1,\ldots , q_n, r_1,\ldots , r_n \geq 0}
 \hskip-0.7cm
 a^k b^l t_1^{p_1}\cdots t_n^{p_n} \re^{q_1 a t_1 + \cdots + q_n a t_n
   +
   r_1 b t_1 + \cdots + r_n b t_n}, 
$$ 
 excluding factorizations and simplifications of expressions.
  
\begin{table}[H]
   \centering 
\begin{tabular}{c|c|c|c|c|c|c|} 
\cline{2-3}
& \multicolumn{2}{c|}{One variable}  
\\ 
\cline{1-5}
  \multicolumn{1}{|c|}{Cumulant} & Time &  Count & \multicolumn{2}{c|}{All variables} 
\\ 
\cline{1-5} 
\multicolumn{1}{|c|}{Sixth } & 64s & 671 & Time & Count 
\\ 
\cline{1-5} 
\multicolumn{1}{|c|}{Fifth } & 11s & 226 &  2403s & 3288 
\\ 
\cline{1-5} 
\multicolumn{1}{|c|}{Fourth } & 1.7s & 81 & 31s & 536  
\\ 
\cline{1-5}
\multicolumn{1}{|c|}{Third } & 0.5s & 35 & 1.6s & 91 
\\ 
\cline{1-5}
\multicolumn{1}{|c|}{Second } & 0.2s & 12 & 0.3s & 14 
\\ 
\cline{1-5}
\multicolumn{1}{|c|}{First} & 0.06s & 4 
\\
\cline{1-3}
\end{tabular} 
\caption{Counts of summands and cumulant computation times in Maple.} 
\label{t1}
\end{table} 

\vskip-0.2cm

 The tables also display the corresponding computation times on a
 8-core laptop computer with 8Gb RAM.
 Symbolic computation appears faster with Maple, although
 computation times become similar at the order six and above.

\noindent 
 Figures~\ref{fig3-a} and \ref{fig3-b} show the
 numerical evaluation of fifth and sixth joint moments. 
 $m_5(t_1,t_2,t_3,t_4,t_5)$ and $m_6(t_1,t_2,t_3,t_4,t_5,t_6)$.
 
\vskip-0.2cm

\vskip-0.2cm

\noindent
 Computation times are presented in seconds for symbolic calculations 
 using the variables $a,b,t_1,\ldots , t_n$,
 and can be significantly shorter when the variables are set to specific 
 numerical values.
 Moment computations times in Table~\ref{t2} are similar to those
 of Table~\ref{t1}, and can be sped up if cumulant functions are memoized
 after repeated calls. 
 
\noindent

\begin{table}[H] 
  \centering
  \begin{tabular}{c|c|c|c|c|c|c|} 
\cline{2-3}
& \multicolumn{2}{c|}{One variable} 
\\ 
\cline{1-5}
 \multicolumn{1}{|c|}{Moment}   & Time &  Count & \multicolumn{2}{c|}{All variables} 
\\ 
\cline{1-5}
\multicolumn{1}{|c|}{Sixth} & 66s & 2159 & Time & Count 
\\ 
\cline{1-5}
\multicolumn{1}{|c|}{Fifth} & 11s & 762 & 2544s & 27116 
\\ 
\cline{1-5} 
\multicolumn{1}{|c|}{Fourth} & 1.9s & 265 & 35s & 2236 
\\ 
\cline{1-5} 
\multicolumn{1}{|c|}{Third} & 0.5s & 88 & 
1.8s & 266 
\\ 
\cline{1-5}
\multicolumn{1}{|c|}{Second} & 0.2s & 22 & 0.5s & 29 
\\ 
\cline{1-5}
\multicolumn{1}{|c|}{First} & 0.05s & 4 
\\
\cline{1-3}
\end{tabular} 
\caption{Counts of summands and moment computation times in Maple.} 
\label{t2}
\end{table} 

\vskip-0.2cm

\noindent
 One-variable examples are computed using the following codes 
 which use Bell polynomials instead of sums over partitions. 

\medskip

\begin{lstlisting}[language=Maple]
kz := proc(z, n, a, b, t) local tmp, z1, k; option remember; 
  if n = 1 then if a = b then return 1 + a*(t - z); else return b/(b - a) + a*exp((a - b)*(t - z))/(a - b); end if; end if;
  tmp := 0; z1 := []; for k from n by -1 to 2 do z1 := [op(z1), kz(y + z, n - k + 1, a, b, t)]; tmp := tmp + IncompleteBellB(n, k, op(z1)); end do;
  return int(a*exp((a - b)*y)*tmp, y = 0 .. t - z); end proc;
c := proc(a, b, t, n) local y, k, z, temp; option remember; temp := 0; y := [];
  for k from n by -1 to 1 do y := [op(y), kz(z, n - k + 1, a, b, t)]; temp := temp + IncompleteBellB(n, k, op(y)); end do;
  return int(temp, z = 0 .. t); end proc;
m := proc(a, b, t, n) local tmp, z, k; option remember; 
tmp := 0; if n = 0 then return 1; end if; z := []; for k from n by -1 to 1 do z := [op(z), c(a, b, t, n - k + 1)]; tmp := tmp + IncompleteBellB(n, k, op(z)); end do;
  return tmp; end proc;
\end{lstlisting}

\vskip-0.6cm

\noindent 
For example, the second moment of $X_t$ is obtained in Maple 
by the command
 ${\rm m}(a,b,t,2)$, which yields 
\begin{align*} 
 \E [ X_t^2 ]  = & 
  \frac{(b^2 t + a (-1 + \re^{(a - b) t} - b t))^2}{(a - b)^4}
  \\
  &
  + \frac{1}{2 (a - b)^4}
  \big(
  \re^{-2 b t} (2 b^4 \re^{2 b t} t
  \\
   &
  + 
  a^2 b (-\re^{2 a t} + \re^{2 b t} + 4 b \re^{(a + b) t} t)
  \\
   & 
   - 
   2 a b^2 (-3 \re^{(a + b) t} + \re^{2 b t} (3 + b t))
   \\
    & + 
    2 a^3 (\re^{2 a t} - \re^{(a + b) t} (1 + 2 b t)))\big). 
\end{align*}
 The same result can be obtained in Mathematica from the command 
 ${\rm m}[a,b,t,2]$ using the code below.
 
\medskip
 
\begin{lstlisting}[language=Mathematica]
kz[z_, n_, a_, b_, t_] := 
  kz[z, n, a, b, t] = (Module[{tmp, k, y, z1}, 
     If[n == 1, If[a === b, Return[1 + a*(t - z)], 
       Return[b/(b - a) + a*E^((a - b)*(t - z))/(a - b)]]]; tmp = 0; 
     z1 = {}; For[k = n, k >= 2, k--, z1 = Append[z1, Block[{i = n - k + 1, u = y + z}, kz[u, i, a, b, t]]]; 
      tmp += BellY[n, k, z1];]; a*Integrate[E^((a - b)*y)*tmp, {y, 0, t - z}]]);
c[a_, b_, t_, n_] := c[a, b, t, n] = (Module[{y, k, z, temp}, temp = 0; y = {}; For[k = n, k >= 1, k--, 
      y = Append[y, Block[{i = n - k + 1, u = z}, kz[u, i, a, b, t]]]; temp += BellY[n, k, y]]; Return[Integrate[temp, {z, 0, t}]]]);
m[a_, b_, t_, n_] := m[a, b, t, n] = (Module[{tmp, z, k}, tmp = 0; If[n == 0, Return[1]]; z = {}; For[k = n, k >= 1, k--, z = Append[z, c[a, b, t, n - k + 1]]; tmp += BellY[n, k, z]]; tmp])
\end{lstlisting}

\vspace{-0.8cm} 

\section*{Appendix A - Maples codes} 

\begin{lstlisting}[language=Maple]
kz := proc(z, a, b, t::list) local pm, pp2, p, pp, tmp, k, y, h, i, j, ii, u, n, zz, c; option remember; n := nops(t);
  if n = 1 then if a = b then return 1 + a*(t[1] - z); else return b/(b - a) + a*exp((a - b)*(t[1] - z))/(a - b); end if; end if;
  tmp := 0; pp2 := Iterator:-SetPartitions(n); for pp in pp2 do p := pp2:-ToSets(pp);
  if 2 <= nops(p) then c := 1; for i to nops(p) do c := c*kz(y + z, a, b, map(op, convert(p[i], list), t)); end do;
  tmp := tmp + c; end if; end do; return a*int(exp((a - b)*y)*tmp, y = 0 .. t[1] - z); end proc;
\end{lstlisting}

\vskip-0.6cm

\begin{lstlisting}[language=Maple]
c := proc(a, b, t::list) option remember; local y, e, k, pm, tmp, p2, pp, p, c, i, zz, j, u, ii, n; n := nops(t);
  tmp := kz(y, a, b, t); if 2 <= n then pm := Iterator:-SetPartitions(n); for pp in pm do p := pm:-ToSets(pp);
  if 2 <= nops(p) then e := 1; for i to nops(p) do e := e*kz(y, a, b, map(op, convert(p[i], list), t)); end do; tmp := tmp + e;
  end if; end do; end if; return int(tmp, y = 0 .. t[1]); end proc;
\end{lstlisting}

\vskip-0.6cm

\begin{lstlisting}[language=Maple]
m := proc(a, b, t::list) option remember; local y, e, k, u, ii, pm, tmp, p2, pp, p, i, zz, j, n; n := nops(t); tmp := c(a, b, t);
  if 2 <= n then pm := Iterator:-SetPartitions(n); for pp in pm do p := pm:-ToSets(pp); if 2 <= nops(p) then e := 1; for i to nops(p) do e := e*c(a, b, map(op, convert(p[i], list), t)); end do; tmp := tmp + e;
  end if; end do; end if; return tmp; end proc;
\end{lstlisting}

\vspace{-1.1cm}

\section*{Appendix B - Mathematica codes}

\begin{lstlisting}[language=Mathematica]
Needs["Combinatorica`"];
kz[z_, a_, b_, t__] := 
  kz[z, a, b, t] = (Module[{tmp, y, i, c, n}, n = Length[t]; 
     If[n == 1, If[a === b, Return[1 + a*(t[[1]] - z)], 
       Return[b/(b - a) + a*E^((a - b)*(t[[1]] - z))/(a - b)]]]; 
     tmp = 0; Do[c = 1; If[Length[p] >= 2, For[i = 1, i <= Length[p], i++, 
        c = c*Block[{u = y + z, v = t[[p[[i]]]]}, kz[u, a, b, v]]]; 
       tmp += c], {p, SetPartitions[n]}]; 
     Return[a*Integrate[E^((a - b)*y)*tmp, {y, 0, t[[1]] - z}]]]);
\end{lstlisting}

\vskip-0.6cm

\begin{lstlisting}[language=Mathematica]
c[a_, b_, t__] := 
  c[a, b, t] = (Module[{y, e, tmp, n, i}, n = Length[t]; tmp = 0; 
     Do[e = 1; For[i = 1, i <= Length[p], i++, 
       e = e*Block[{u = y, v = t[[p[[i]]]]}, kz[u, a, b, v]]]; 
       tmp += Flatten[{e}][[1]],
       {p, SetPartitions[n]}]; 
     Return[Integrate[tmp, {y, 0, t[[1]]}]]]);
\end{lstlisting}
\vskip-0.6cm

\begin{lstlisting}[language=Mathematica]
m[a_, b_, t__] := (Module[{n, e, i, tmp}, tmp = 0; n = Length[t]; If[n == 0, Return[1]];
   Do[e = 1; For[i = 1, i <= Length[p], i++, e = e*c[a, b, t[[p[[i]]]]]]; 
    tmp += e, {p, SetPartitions[n]}]; Flatten[{tmp}][[1]]])
\end{lstlisting}

\vspace{-1.1cm}

\footnotesize 

\def\cprime{$'$} \def\polhk#1{\setbox0=\hbox{#1}{\ooalign{\hidewidth
  \lower1.5ex\hbox{`}\hidewidth\crcr\unhbox0}}}
  \def\polhk#1{\setbox0=\hbox{#1}{\ooalign{\hidewidth
  \lower1.5ex\hbox{`}\hidewidth\crcr\unhbox0}}} \def\cprime{$'$}

\end{document}